# Real-time coherent diffraction inversion using deep generative networks

Mathew J. Cherukara,[1*] Youssef S.G. Nashed[2] and Ross J. Harder[1]

[1]*Advanced Photon Source, Argonne National Laboratory, Argonne, IL, 60439*

[2]*Mathematics and Computer Science, Argonne National Laboratory, Argonne, IL, 60439*

**Phase retrieval, or the process of recovering phase information in reciprocal space to reconstruct images from measured intensity alone, is the underlying basis to a variety of imaging applications including coherent diffraction imaging (CDI). Typical phase retrieval algorithms are iterative in nature, and hence, are time-consuming and computationally expensive, precluding real-time imaging. Furthermore, iterative phase retrieval algorithms struggle to converge to the correct solution especially in the presence of strong phase structures. In this work, we demonstrate the training and testing of CDI NN, a pair of deep deconvolutional networks trained to predict structure and phase in real space of a 2D object from its corresponding far-field diffraction intensities alone. Once trained, CDI NN can invert a diffraction pattern to an image within a few milliseconds of compute time on a standard desktop machine, opening the door to real-time imaging.**

## INTRODUCTION

Central to many imaging techniques including coherent X-ray diffraction imaging,[1,2] electron microscopy,[3] astronomy[4] and super-resolution optical imaging,[5] is the process of phase retrieval, or the recovery of phase information from the measured intensities alone. In particular, in X-ray coherent diffraction imaging (CDI), an object is illuminated with a coherent X-ray beam and the resulting far-field diffraction pattern is measured. This far-field diffraction pattern is the modulus of the Fourier transform of the object, and phase retrieval algorithms are used to reconstruct the measured object by recovering the lost phase information. As such, the imaging methods are extremely sensitive to any material properties that contribute a phase to the scattered beam.[6] In particular, when measured in the vicinity of a Bragg peak, the measured coherent far-field X-ray diffraction pattern encodes the strain within the object in the local asymmetry of the coherent diffraction pattern around the Bragg peak.[7] The strain induces distortion in the lattice, which manifests itself in the scattered beam as an additional phase. Upon successful inversion of the diffraction pattern to an image, the local distortion of the crystal lattice is then displayed as a phase in the complex image of the sample.[8] X-ray CDI and Bragg CDI (BCDI) in particular have been widely used to provide an unique 4D view of dynamic processes including phonon transport,[9,10] transient melting,[11] dissolution and recrystallization,[12] phase transformations,[13] grain growth[14] and device characterization.[15,16] Notwithstanding its widespread use, reciprocal space phase retrieval algorithms suffer from several shortcomings. Firstly, the iterative phase retrieval algorithms that are commonly used, such as error-reduction (ER) and hybrid input-output (HIO)[17] or difference map (DM)[18] are time consuming, requiring thousands of iterations and multiple random initializations to converge to a solution with high confidence.[19] Even when converged, there is no guarantee that the converged solution corresponds to the global minimum or optimum solution. Furthermore, such algorithms often fail to converge in presence of strong phase structures for instance those associated with multiple defects in materials.[20] Additionally, algorithmic convergence is often sensitive to certain iterative phase retrieval parameters such as initial guesses, shrinkwrap threshold,[21] choice of algorithms and their combinations.[22] Finally, a necessary mathematical condition for reciprocal space phase retrieval is that the measured intensities are oversampled by at least a factor of two.[23] In practice, this requirement translates to necessitating a minimum of two pixels per coherent feature on the detector. Consequently, this limits the extent of reciprocal space that is accessible for a given detector size and x-ray wavelength.

---

* mcherukara@aps.anl.gov

Neural networks have been described as universal approximators, with the ability to represent a wide variety of functions.[24] As such, they have been used for an enormous variety of applications ranging from natural language processing and computer vision to self-driving cars.[25] More recent work has predominantly involved the use of deep neural networks, so termed because of the manner in which they are structured to learn increasingly more complex features or hierarchal representations with successive layer of neurons.[26] In particular, deep deconvolutional networks have found a variety of applications in various imaging techniques, ranging from automated image segmentation of electron microscopy images,[27] image reconstruction from magnetic resonance imaging (MRI),[28] to the enhancement of images from mobile phone microscopes.[29]

Specific to the problem of phase recovery, deep neural networks have been used in holographic image reconstruction,[30] phase retrieval following spatial light modulation (SLM),[31] optical tomography,[32] and as a denoiser in iterative phase retrieval.[33] We note that none of these works represent an end-to-end solution to the far-field reciprocal space phase retrieval problem.

In this work, we train two deep deconvolutional networks to learn the mapping between 2D coherent diffraction patterns (which are the magnitudes of an object's Fourier transform) and the corresponding real-space structure and phase. While we have trained these neural networks (NNs) with the intention of applying them to CDI measurements, in particular to BCDI measurements, the approach outlined in this study is easily transferrable to any imaging modality that requires reciprocal space phase retrieval. Once trained, these deconvolutional networks, which we term CDI NN can predict the structure and phase of test data within a few milliseconds on a standard desktop machine. This is thousands of times faster than what is achievable with iterative phase retrieval algorithms currently in use. Such real-time image reconstruction has the potential to revolutionize the various advanced imaging modalities that rely on phase retrieval and is essential to performing *in-situ* and *operando* characterization studies of rapidly evolving samples for experimental feedback.

**RESULTS**

**Coherent Diffraction Imaging (CDI)**

BCDI measurements are typically performed on compact objects such as isolated nanoparticles or single grains within a polycrystalline material. To simulate the compact structure associated with an isolated particle or a single grain, we use convex polygons of random size and shape. Points within these polygons are complex values with magnitude of 1, while points lying outside have a magnitude of 0. We also give the edges of the polygons a Gaussian transition from 1 to zero with a width of one pixel. Points within the polygon have a spatially-varying complex phase that simulates the distortion of a crystalline lattice due to strain within a material. Without loss of generality, this phase can represent any structural inhomogeneity that modifies the phase of a scattered beam. Finally, to obtain the diffraction signal corresponding to the object, we take the magnitudes of the two-dimensional (2D) Fourier transform (FT) of the complex valued compact object.

**CDI NN's structure and training**

Our generative network, CDI NN is a feed-forward network consisting of two parts, as shown in Figure 1. The first part is a convolutional autoencoder that is responsible for finding a representation, or encoding, of an input image in feature space. This encoding is then propagated through a deconvolutional decoder to generate an output image. The overall network is trained in a supervised fashion, where the output image is known a priori.[34,35] We train two networks with identical architecture, one that takes diffraction amplitudes as input and produces object shape as output (structure CDI NN or sCDI NN), and a second that takes diffraction amplitudes as input while outputting real space phase information (phase CDI NN or pCDI NN). The convolutional and maxpooling operations serve to transform the image data (in this case the diffracted amplitudes) into feature space, while the deconvolutional and upsampling operations serve to transform back from feature space into pixel space.

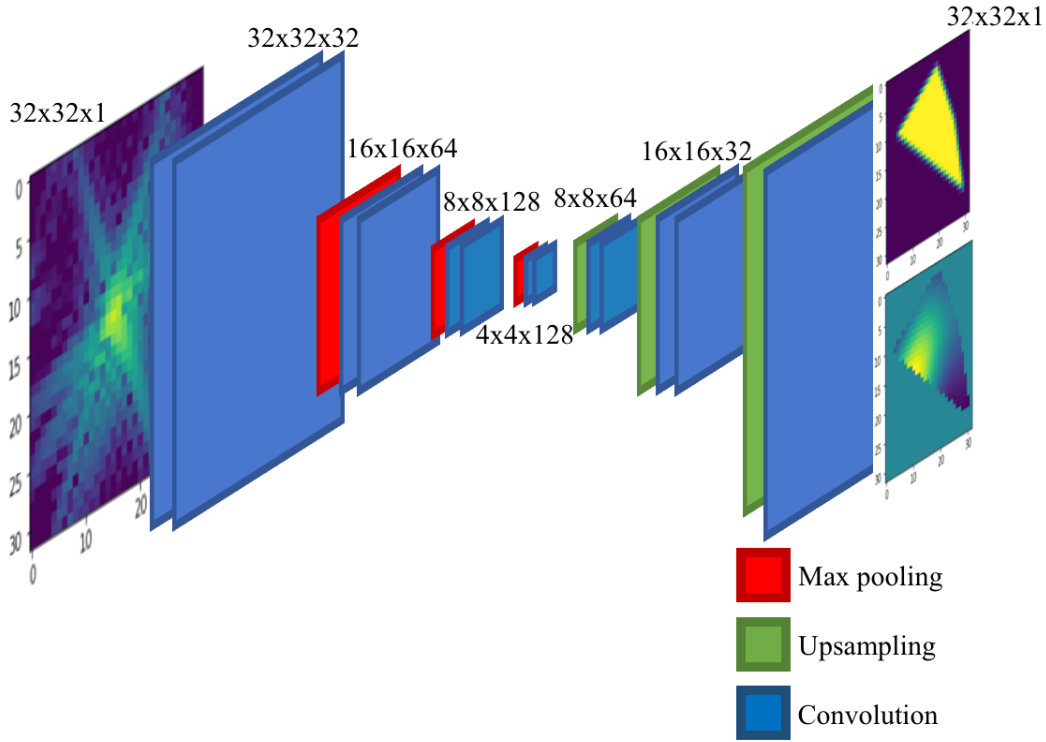

*Figure 1: Structure of the deep generative network CDI NN. CDI NN is implemented using an architecture composed entirely of convolutional, max pooling and upsampling layers. All activations are rectified linear units (ReLU) except for the final convolutional layer which uses sigmoidal activations.*

To train the two networks that together compose CDI NN (sCDI NN and pCDI NN), we exposed the networks to 180,000 training examples consisting of diffraction magnitudes and the corresponding real space structure and phase. Each instance of the training data was generated as described in **methods.** We set aside 20,000 instances from the generated training data for model validation at the end of each training pass (also see **methods**). Figure 2 A shows the training and validation loss as a function of epochs for sCDI NN, while fig.2 B shows the training and validation loss as a function of epoch for pCDI NN. For both networks, we see that the weights converge within 10 epochs as evinced by the behavior of the validation loss. We found that training for more epochs causes the validation loss to diverge, suggesting that the network was beginning to overfit to training data beyond 10 training epochs.

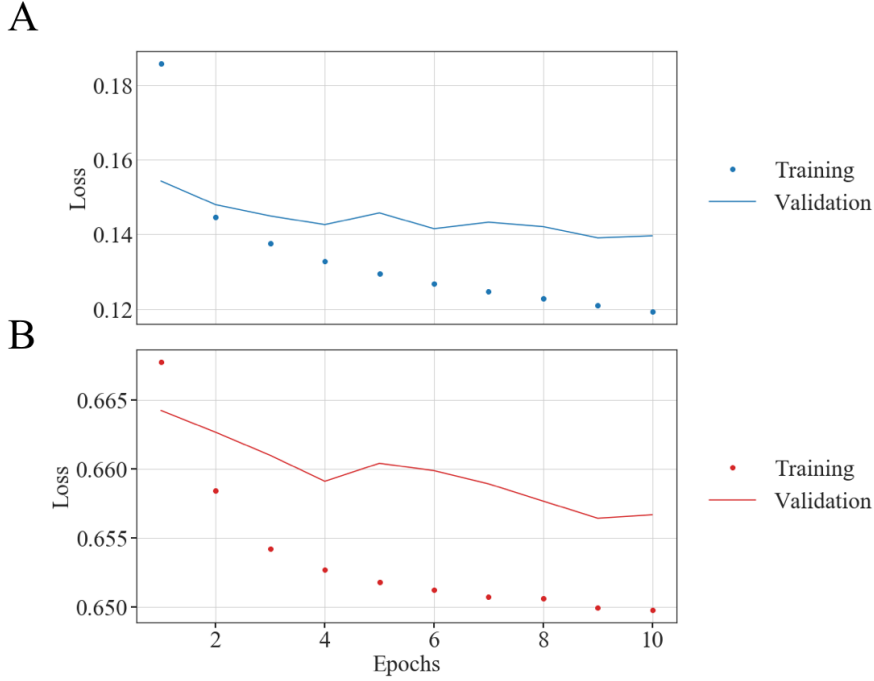

*Figure 2:* **Training and validation.** *Training and validation loss as function of training epoch for A sCDI NN and B pCDI NN. Training was stopped after 10 epochs beyond which the validation loss diverged, suggesting overfitting beyond 10 epochs.*

**CDI NN's performance on test data**

To test the performance of the trained CDI NN networks, we evaluate their performance on a new set of 1000 test cases that was not shown to the networks at any point during training. In testing, we import the trained neural networks' topology and optimized weights and evaluate its ability to reconstruct real space structure and phase from input diffraction patterns. Figure 3 shows random samples of the performance of the network in testing. The first row (fig. 3 A) shows the input diffracted amplitudes. Fig.3 B (second row) shows the corresponding ground truth objects, while Fig.3 C (third row) shows the structures predicted by sCDI NN. We observe an excellent match between the predicted and true object structures. Figure 3D shows the true phase structure, while fig. 3E shows the phases predicted by pCDI NN. Again, we observe a good agreement between the prediction of CDI NN and the actual phase structure. We note that the images shown in fig. 3E are bounded by the structure predicted by sCDI NN in fig. 3C, i.e, phases outside of the predicted object shape are set to 0. We use a threshold of 0.1 to define the boundary of the object. We also draw the reader's attention to the third example, where we note that the actual (B) and predicted (C) objects are twin images of each other, and that they can be obtained from each other through a centrosymmetric inversion and complex conjugate operation. Both images are equivalent solutions to the input diffraction pattern. We observe several such instances where shape predicted by CDI NN is the twin image, especially when the phase structure is weak or constant and the corresponding diffraction image is nearly centrosymmetric.

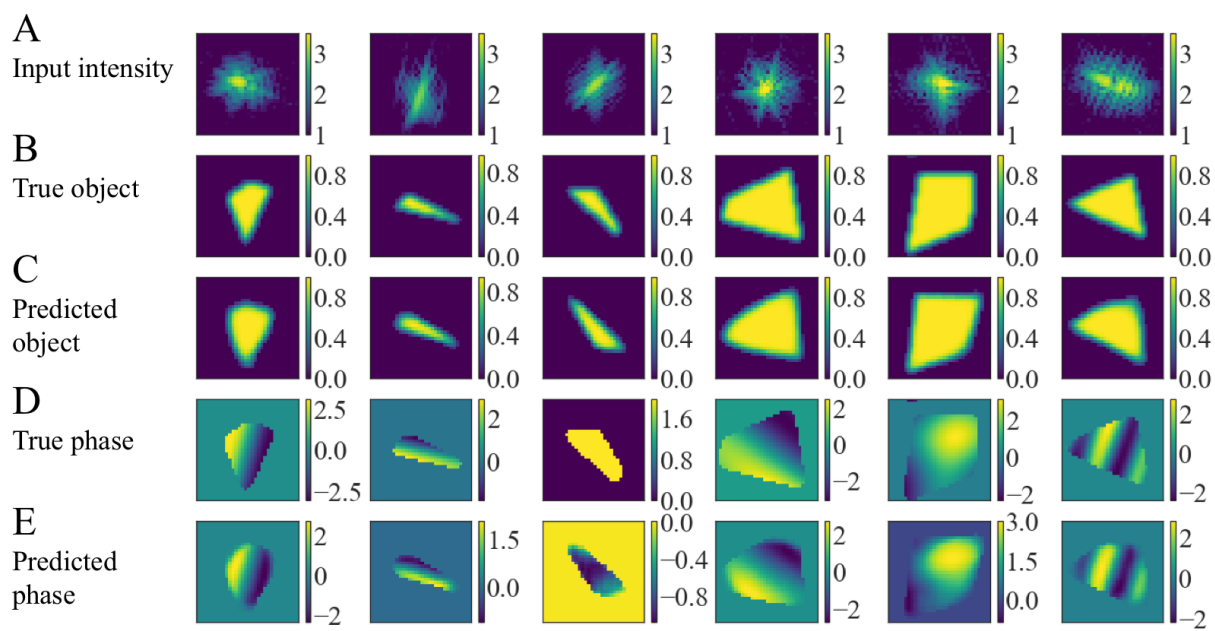

*Figure 3: Examples of the performance of CDI NN in testing.* *A input diffraction intensities to CDI NN. B Actual shape of the corresponding object and C the shape predicted by sCDI NN. D the actual phase structure of the object and E, the phase structure predicted by pCDI NN. CDI NN successfully recovers structure and shape for a variety of different shapes and phase structures.*

## DISCUSSION

### The best and brightest and the worst and dullest

To gain some insight into the strengths and weaknesses of CDI NN in predicting an object's structure and phase from its diffraction pattern, we quantify the error in each prediction by comparing the prediction to the ground truth. We make the comparison in reciprocal space, where we take the retrieved image (structure and phase), compute the reciprocal space amplitudes and compare those with the amplitudes given to CDI NN.

In Figure 4A we plot a histogram of the $\chi^2$ error for each of the test cases. To compute this error for each of the test cases, we take the predictions from sCDI NN and pCDI NN and compute the FT to obtain the predicted diffraction intensity. The error $\chi^2$ is then given by:

$$\chi^2 = \frac{\sum_{i=1}^{N}\left(\sqrt{I_p^i}-\sqrt{I_t^i}\right)^2}{\sum_{i=1}^{N} I_t^i} \tag{1}$$

where $I_t^i$ are the true diffraction intensities and $I_p^i$ are the predicted diffraction intensities at each pixel. Figure 4 B shows a zoom of the histogram at the lowest $\chi^2$ error, i.e, where the predictions are the best. The panels on the right show the 5 best predictions as computed by the error metric. Fig. 4 C shows the input diffraction intensity, fig. 4 E the true object structure, fig. 4 F the predicted shape, fig. 4 G the actual phase structure and fig. 4 H the predicted phase structure. Figure 4 D shows the predicted diffraction intensity which is obtained by taking an FT of the predicted shape and predicted phase. We observe that the best predictions as defined by the error in diffraction intensities, is found when the objects are large, with relatively weak phase structure. This is perhaps unsurprising since in these situations, the diffraction pattern is quite symmetric with most of the intensity centered around the Bragg peak.

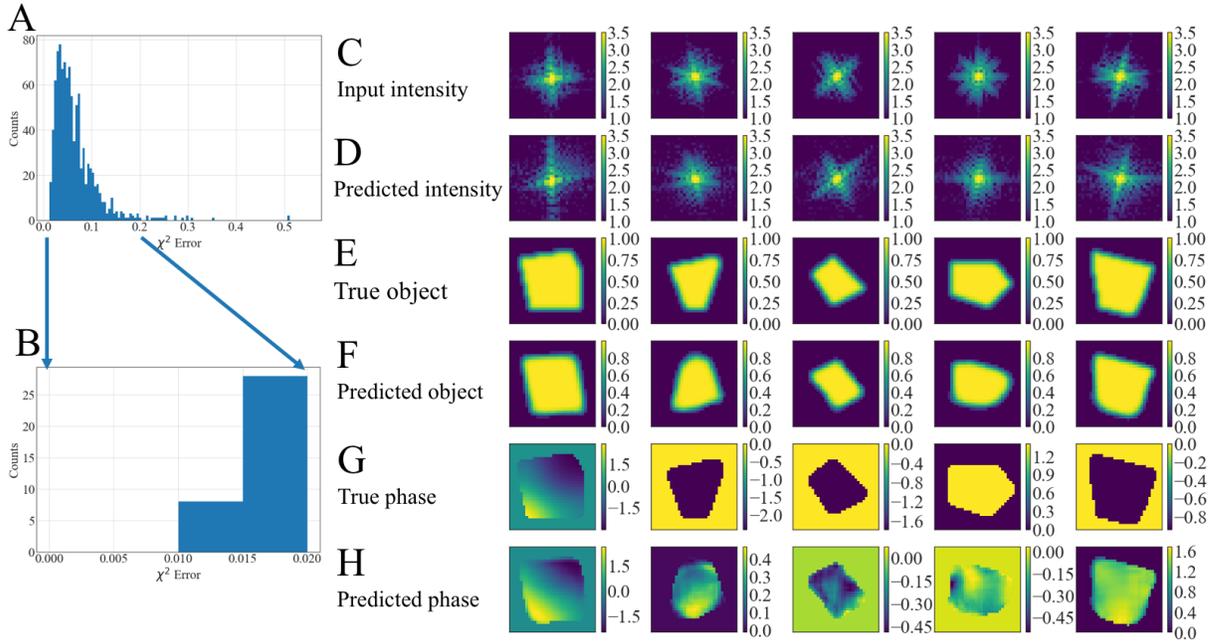

*Figure 4: Best predictions of CDI NN. A Histogram of the error computed for the test samples. B Close-up view of the predictions with the lowest $\chi^2$ error. C input diffraction patterns, D predicted intensities computed by taking an FT of the complex object obtained from the predicted shape and phase. E,F corresponding actual shape and predicted shape. G, H corresponding actual phase and predicted phase. CDI NN fares well when faced with large objects with weak phase.*

Conversely, Figure 5 shows the 5 worst predictions as inferred by the $\chi^2$ error metric. Fig. 5 B shows a zoomed in view of the histogram showing the error of the worst cases, with the maximum computed error being $\chi^2 \sim 0.5$. Here, we note that in contrast to the best predictions, CDI NN fares the worst when faced with smaller objects and strong phase structures (Fig.5 E-H). In such instances, the Bragg peak center is poorly defined, and the diffracted intensity is spread across a large range of q as seen in Fig. 5 C-D. However, we see that even for the most difficult instances, the predicted shape and phase structures are reasonable. One would expect to see such strong phase structure in the presence of extended defects such as stacking faults or twin planes, underscoring the strength of our deep learning approach to image reconstruction.[14]

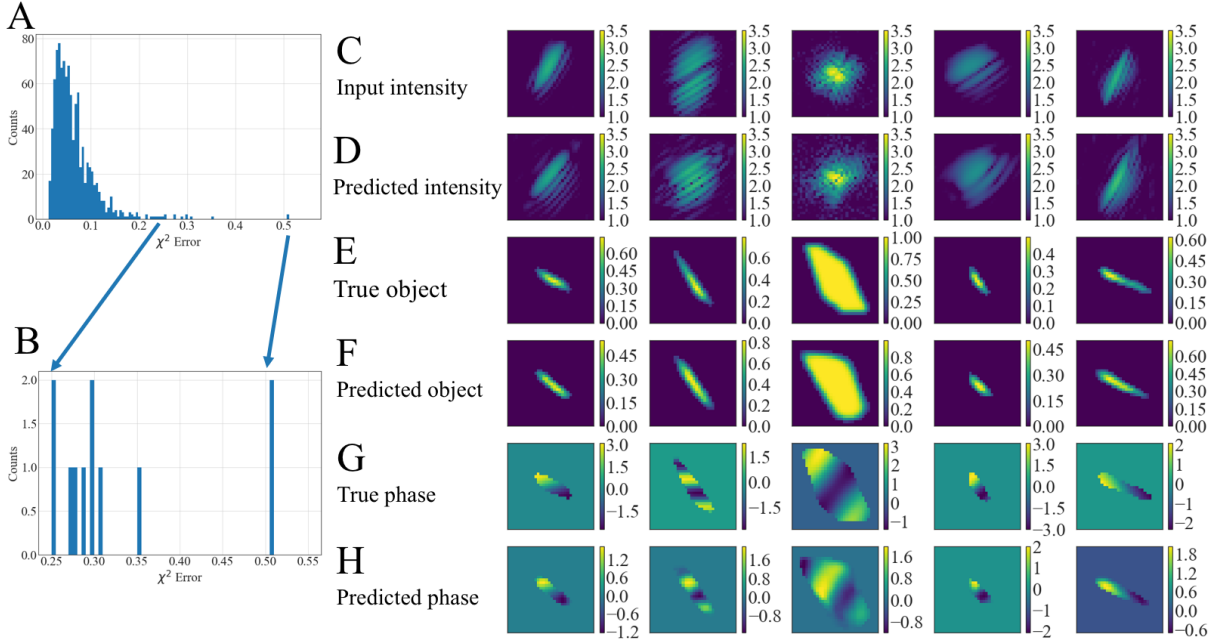

*Figure 4: Worst predictions of CDI NN.* A Histogram of the error computed for the test samples. B Close-up view of the predictions with the highest $\chi^2$ error. C input diffraction patterns, D predicted intensities computed by taking an FT of the complex object obtained from the predicted shape and phase. E,F corresponding actual shape and predicted shape. G, H corresponding actual phase and predicted phase. CDI NN performs worst when faced with diffraction patterns corresponding to small objects with strong phase.

**Network activation maps**

To investigate the nature of the features that the convolutional layers learn, we study the layer activations for different input diffraction patterns. Figure 6 shows average activation maps of the 2$^{nd}$ convolutional layer and 4$^{th}$ convolutional layer (see Fig. 1) for 3 different input diffraction patterns (Fig. 6 A). These activation maps represent the average activations of the 32 convolutional filters that make up the 2$^{nd}$ convolutional layer (Fig. 6 B, D), and an average of the 64 convolutional filters that make up the 4$^{th}$ convolutional layer (Fig. 6 C, E) for the two networks (sCDI NN and pCDI NN). We note that the images shown in Fig. 6 C, E have been interpolated from 16x16 to 32x32 to enable a straightforward comparison. For both networks, we observe that at the 2nd convolutional layer, the network focusses on regions close to the brightest pixels at the center of the Bragg peak. Interestingly, we observe differing behaviors of the two networks at the 4th convolutional layer. At the 4th convolutional layer, the structure network (sCDI NN) focusses solely on regions at higher scattering vector (higher Q) relative to the center of the diffraction pattern, i.e. choosing to focus on finer scale features. In slight contrast, the phase network (pCDI NN), continues to focus strongly on the center of the diffraction pattern, while also paying more attention to pixels at higher Q. In both networks, successive layers start paying attention to data at higher Q (finer spatial resolution data), and this suggests that CDI NN progressively learns higher order features in the image in encoding the structure of the image.

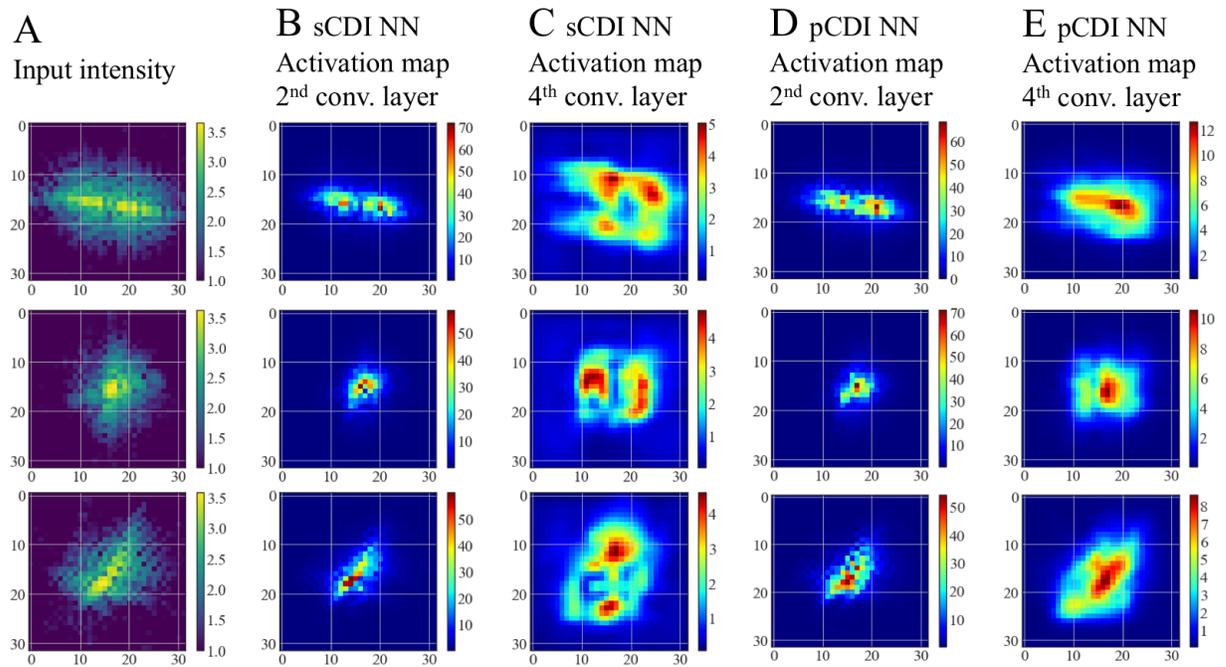

*Figure 6: Activation maps of select convolutional layers.* *A input diffraction patterns. B average activation of the 2nd convolutional layer of sCDI NN. C average activation of the 4th convolutional layer of sCDI NN. D average activation of the 2nd convolutional layer of pCDI NN. E average activation of the 4th convolutional layer of pCDI NN. Successive layers of CDI NN learn features at higher Q (higher resolution).*

**Conclusion**

In conclusion, to the best of our knowledge, this is work is the first demonstration of an end-to-end deep learning solution to the phase retrieval problem in the far-field. We believe the results described in this manuscript have widespread ramifications for both BCDI experiments of the future for which this study was designed as well as other imaging modalities reliant on successful phase retrieval.

CDI NN is thousands of times faster than traditional phase retrieval and requires only modest resources to run. We note that while the training of CDI NN was performed on a dual GPU machine (~1 hour training time); once trained, CDI NN can easily be deployed at a standard desktop at the experiment's location. Indeed, the test cases in this manuscript were run on the CPUs of a 2013 Mac Pro desktop with Quad-Core Intel Xeon E5, where the prediction time was ~2.7 milliseconds. We expect that such real-time feedback will be crucial to coherent imaging experiments especially in the light of ongoing upgrades to major light sources such as Advanced Photon Source Upgrade project, Linac Coherent Light Source 2 and National Synchrotron Light Source II. Additionally, CDI NN was shown to be successful at recovering structure and phase even in the presence of strong phase structures that heavily distorts the coherent diffraction pattern about a Bragg peak, and this shows strong promise for the successful reconstruction of objects that have a high density of defects.

Finally, whereas oversampling is a necessary condition for phase retrieval algorithms to work, CDI NN does not require oversampled data. Experimentally, the relaxation of this oversampling requirement will translate to several further advantages. For a given detector configuration (pixel size and distance), higher energies (that allow deeper penetration into material) and access to a higher volume in reciprocal space (that provides increased resolution) will become possible.

# METHODS

## Simulated training data:

Each instance of the training data was generated as follows; first a random convex object was created from a convex hull of a random scattering of points within a 32x32 grid. Array values within the object were set to 1, while values outside are set to 0. A Gaussian blur 1 pixel in width was applied to the object to smooth the edges. A second array (also 32x32) was created with a random, spatially varying phase field. For convenience, phases outside the compact object were set to 0. The corresponding diffraction pattern was then generated by taking the FT of a complex valued array created from the object's amplitudes and phases. Only the amplitude information from the computed diffraction patterns was retained for both training and testing of CDI NN.

## CDI NN training:

Training was performed in parallel on two NVIDIA K40 GPUs using the Keras package running the Tensorflow backend.[36,37] We trained the networks for 10 epochs each using a batch size of 256. The training for each network took less than half an hour when trained in parallel across the two GPUs. At each step, we used adaptive moment estimation (ADAM)[38] to update the weights while minimizing the per-pixel loss as defined by the crossentropy. We computed the performance of the network at the end of each training epoch using the validation set.

## Data Availability

The trained network, test data and accompanying Jupyter notebooks of Python code are available upon reasonable request from the corresponding author.

**Acknowledgements**

This work was supported by Argonne LDRD 2018-019-N0 (A.I C.D.I: Atomistically Informed Coherent Diffraction Imaging). An award of computer time was provided by the Innovative and Novel Computational Impact on Theory and Experiment (INCITE) program. This work also used computational resources at the Advanced Photon Source. The Advanced Photon Source and the Argonne Leadership Computing Facility are supported by the U.S Department of Energy, Office of Science, Office of Basic Energy Sciences, under Contract No. DE-AC02-06CH11357

**Author contributions**

M.J.C., Y.N and R.H designed the research. M.J.C built, trained and tested the deep learning networks. All authors contributed to the analysis, discussion and writing of the manuscript.

**Competing interests**

The authors declare no competing interests.